\documentclass[sigconf,,authorversion,nonacm]{acmart}

\setcopyright{none}
\renewcommand\footnotetextcopyrightpermission[1]{}
\setcopyright{none}
\makeatletter
\renewcommand\@formatdoi[1]{\ignorespaces}
\makeatother

\usepackage{multicol}

\AtBeginDocument{%
   \providecommand\BibTeX{{%
     \normalfont B\kern-0.5em{\scshape i\kern-0.25em b}\kern-0.8em\TeX}}}

\begin{document}

\title{Detection of Fake Users in SMPs Using NLP and Graph Embeddings}


%
\author{Manojit Chakraborty}
\affiliation{%
  \institution{Robert Bosch Research and Technology Center, India}
  \city{Bangalore}
  \country{India}
}

\author{Shubham Das}
\affiliation{%
  \institution{Swiggy}
  \city{Bangalore}
  \country{India}
}

\author{Radhika Mamidi}
\affiliation{%
  \institution{IIIT Hyderabad}
  \city{Hyderabad}
  \country{India}
}
\begin{abstract}
  Social Media Platforms (SMPs) like Facebook, Twitter, Instagram etc. have large user base all around the world that generates huge amount of data every second. This includes a lot of posts by fake and spam users, typically used by many organisations around the globe to have competitive edge over others. In this work, we aim at detecting such user accounts in Twitter using a novel approach. We show how to distinguish between \textit{Genuine} and \textit{Spam} accounts in Twitter using a combination of Graph Representation Learning and Natural Language Processing techniques.
\end{abstract}


\maketitle

\section{Introduction}

Fake profiles are using Social Media Platforms and making them unreliable for people. They are being massively used to get information illegitimately, steal identity, defame someone, spread misinformation and malware, and boost follower counts for popularity and reach. It is accompanied by online spamming where machine learning models are extensively used to put up fake reviews for products or services. Only on Twitter, 3 out of every 20 accounts are automated bots and 1 out of every 21 tweets is spam \cite{dutse}, which simply raises question on user's true identity and validity of the posted content. A large number of attempts are being made in order to detect fake account on social media platforms. However, at the same time, spamming techniques are evolving rapidly to break through such detection systems, which makes many existing solutions ineffective against these new techniques.

\section{Related Works}
Igawa et. al\cite{igawa} presented a pattern recognition based approach which used weighing mechanism called as Lexicon Based Coefficient Attenuation. Random Forest Classifier and Multilayer Perceptron are used for human,non-human classification and human,bot and Cyborg classification.
Varol et. al \cite{varol} used public data to extract around a thousand features about users: followers, followings, interactions and other activities and identified different types of accounts, including self promotions, spamming and fake news spreading accounts.
Dutse et.al \cite{dutse} proposed the use of features independent of historical tweets on Twitter, since they are only available for short time. Features related to each account their pairwise connections to each other was taken into consideration.
Khaled et al. \cite{khaled} proposed the use of SVM-NN which uses a combination of Support Vector Machine and Neural Networks to detect fake/bot accounts on Twitter.
Daouadi et al. \cite{daudi} used deep learning methods on features based on the amount of interaction to and from each Twitter account along with other set of features used previously, for fake user detection.
Abu-El-Rub and Mueen \cite{abu} used trending hashtags to detect bots interested in political trends. Graph based techniques are used to cluster the collected bots and those are fed to supervised learning to detect user's agreement/disagreement to a sentiment and hence identify bots used in political campaigns.

\section{Our Work}
We propose a novel approach of classifying Twitter user accounts into \textit{Genuine} and \textit{Spam}, by using a combination of features engineered from User Metadata, attributes created using various Natural Language Processing techniques, graph centrality values and graph embeddings generated from the followers-followings graph created from our Twitter database.  This database is also a new and updated database of Twitter users and User metadata, generated by us. 

The following section contains the detailed methodology of our dataset preparation, data pre-processing, various feature generation methods using NLP and graph frameworks, selection of features, Machine Learning model selection, model training, evaluation and results from our method.

\subsection{Dataset Preparation}
There are many public datasets on bot/spam account detection related to Twitter, used by the works referred in Section 2. The main problem with those datasets are:
\begin{itemize}
    \item Most of the labelled accounts are suspended/deactivated by Twitter recently and majority of such accounts belonged to the category of fake account.
    \item Features of such accounts were collected a couple of years ago. So, with rapidly changing social media, such features will not reflect current trends which can be crucial in distinguishing fake accounts,
    \item Such datasets do not have tweets or social connections (like followers and followings) list which we could leverage to create social-graph.
\end{itemize}
We use datasets like Midterm-2018-candidates \footnote{https://github.com/vegetable68/Midterm-2018-candidates}, Gilani-2017 \cite{gilani}, Varol-2017 \footnote{https://botometer.osome.iu.edu/bot-repository/datasets/varol-2017/varol-2017.dat.gz} and Twitter API to accumulate a total of 126503 unique accounts. Then these account/Twitter IDs were queried against the Twitter API to detect inactive accounts. After removing such accounts, the dataset had 46354 remaining account IDs along with their class label of \textit{Spam} or \textit{Genuine} account.
Now we use these IDs to get user metadata with the following 37 account features from the Twitter API:

\begin{multicols}{2}
\begin{itemize}
    \item id
    \item user\textunderscore name
    \item screen\textunderscore name
    \item statuses\textunderscore count
    \item followers\textunderscore count
    \item friends\textunderscore count
     \item favourites\textunderscore count
     \item listed\textunderscore count
     \item url
     \item lang

     \item time\textunderscore zone
     \item location
     \item default\textunderscore profile
     \item default\textunderscore profile\textunderscore image
     \item geo\textunderscore enabled 
     \item profile\textunderscore image\textunderscore url 
     \item profile\textunderscore use\textunderscore background\textunderscore image
     \item profile\textunderscore background\textunderscore image\textunderscore url
     \item profile\textunderscore text\textunderscore color
     \item profile\textunderscore image\textunderscore url 
     \item profile\textunderscore sidebar\textunderscore border\textunderscore color
     \item profile\textunderscore background\textunderscore tile 
     \item profile\textunderscore sidebar\textunderscore fill\textunderscore color
     \item profile\textunderscore background\textunderscore image\textunderscore url 
     \item profile\textunderscore background\textunderscore color
     \item profile\textunderscore link\textunderscore color
     \item utc\textunderscore offset
     \item is\textunderscore translator
     \item follow\textunderscore request\textunderscore sent
     \item protected 
     \item verified
     \item notifications
     \item description
     \item contributors\textunderscore enabled
     \item following
     \item created\textunderscore at
\end{itemize}
\end{multicols}

The followers and followings Twitter IDs of each of these accounts were extracted to create an edgelist of $ID1$ and $ID2$ where $ID1$ is the follower ID and $ID2$ is the followed Twitter account ID.
We also use the Twitter API to extract full text of all the tweets and  and retweet counts of each user ID. 

\subsection{Feature Generation}
\subsubsection{\textbf{Social Graph Features}}
Each Twitter account has follower and following list. We created a graph with users as nodes and followers and followings as edges. From this graph we derived various centrality features for each user:
\begin{itemize}
    \item \textbf{Degree Centrality \cite{freeman}}
    
    \item \textbf{Shortest-path Betweenness Centrality \cite{brandes}}
    
    \item \textbf{In-Eigenvector Centrality \cite{newman}}
    \item \textbf{Out-Eigenvector Centrality \cite{newman}}

    \item \textbf{Pagerank Centrality \cite{page}}
    
\end{itemize}
\vspace{2mm}
    \textbf{Node2Vec Embeddings}:
    \newline
    Twitter is a huge social media network. We wanted to leverage the local neighborhood information encoded in the graphical structure of the follower-following graph of Twitter user accounts. For that, a linear structure is needed for each node. Now, Node2Vec \cite{grover} is an algorithmic framework for learning continuous feature representations or embedding vectors for each node in a graph. It aims to preserve network neighborhoods of nodes by creating representation vectors where neighborhood nodes have small euclidean distance between each other.
    \[ node2vec(G(V,E))\>\Rightarrow \mathbb{R}^n\]

    Embedding vector of length 100 ($V_i$) is generated for each node/Twitter ID based on the following/follower graph using node2vec algorithm.
    We ran the Node2Vec Algorithm with the following parameters -
    \begin{itemize}
        \item Number of dimensions - 100
        \item Length of walk per source - 25
        \item Number of walks per source - 10
        \item Return hyperparameter - 0.3
    \end{itemize}
    
\subsubsection{\textbf{Features derived from metadata}}
From Twitter user metadata features, we engineer a set of new features, which can be distinguishing factors between \textit{Genuine} and \textit{Spam} user profiles. \cite{dutse}
\begin{itemize}
    \item \textbf{friends\_count/followers\_count ratio}:
    To analyze the strategies of follower-following, we have the number of followers
(users that are following the current user) and the number of followings (users that the
current user is following) for each account. We calculate -
\begin{center}
    \[Ratio\>=\>\frac{\#\>of\>followings\>/\>friends}{\#\>of\>followers}\]
\end{center}

    \item \textbf{Similarity in Username \& Screen name}:
    Here we calculate the similarity in Username and screen name of the account using Ratcliff and Obershelp gestalt pattern matching algorithm. \footnote{https://docs.python.org/3/library/difflib.html}
    It calculates the ratio as: \textbf{Ratio = 2.0*M/T}, where M = no. of matches , T = total number of elements in both sequences..
    \item \textbf{Account Age}:
    Calculated from the created\textunderscore at field in user metadata and converted to days.
    \item \textbf{Status Count/Account Age}:
    This is a measurement of how active the account is, irrespective of the account age.
    \begin{center}
        \[Activity\>Ratio = \frac{\#\>of\>Statuses}{Account\>age\>in\>days}\]
    \end{center}
    \item \textbf{Favorites Count / Status Count}
    \item \textbf{Shannon Entropy / Profile Description Length}:
\newline
    This is very useful to understand the randomness of the tweet texts for each user.
    Shannon entropy \cite{shannon} of the profile description text is calculated and is divided by the profile description length as:
    \begin{center}
        \[Ratio\>=\>\frac{Shannon\>entropy\>of\>profile\> description}{Length\>of\>profile\>description}\]
    \end{center}
    
\end{itemize}

\subsubsection{\textbf{NLP Based Features}}
From the tweets, we derived several attributes using natural language processing techniques for each user account :
\begin{itemize}
    \item \textbf{Tweet Similarity}:
    It is used to find similarity between two different text corpora. Here, We have used cosine similarity which calculates similarity between text tweets by measuring cosine of the angle between two tf-idf vectors representing the tweets of an user.
    We take top 200 cosine similarity values of pairwise tweet vectors and calculate average of it to get a distinct similarity value for each Twitter user account.
    \item \textbf{Lexical Diversity}:  
    Lexical diversity refers to the vocabulary richness in a given text. Type Token Ratio (TTR) measures richness of a lexicon in a document. Understanding the use of distinct words in various parts of the data corpus is very important for detecting \textit{Spam} accounts.
    \textit{Spam} users typically have low lexical diversity and sophistication compared to \textit{Genuine} users. Legitimate users are expected to use rich and diverse lexicons in tweets depending on the discussion topic. \textit{Spam} users focus on specific targets such as promoting a certain product or marketing to increase the number of their followers. The TTR index is the ratio between the number of ‘types’ (unique word forms) and ‘tokens’ (total number of words) in a text.

    \item \textbf{User Mention Count}:
    User mentions are detected using anything followed by @<Name>
    \item \textbf{Hashtag Count}:
    Hashtag count has been detected using ‘\#’ to show something is trending as followed generally in Twitter.
    We go through each tweet, split them by space and then counting each word which begins with ‘\#’ or ‘@’ and filter them using a special regex to remove misused words.
    \item \textbf{Co-occurring Word Distribution}:
    We obtain a large list of known n-gram spam words from \cite{dutse}. Then we search for these representative spam words in tweets of each user using 2 frequency based approaches - unigram and bigram, to get spam words for individual user. The user ids with occurrences of high frequency spam words show that those Twitter accounts are more likely to be fake or \textit{spam} users.
    
\end{itemize}

\subsection{Feature Selection}

In total, we have 54 features from Twitter user metadata, NLP based features and graph centrality values. Now, two different feature selection methods are used to select most relevant features amongst these, which would be included in the final database to train our detection model. 

\subsubsection{\textbf{SHAP ( SHapley Additive exPlanations  ) \cite{lundberg}}-\textit{Using XGBoost and LightGBM - }}

    The SHAP explanation method computes Shapley values from coalitional game theory. The feature values of a data instance act as players in a coalition. Shapley values tell us how to fairly distribute the “payout” (= the prediction) among the features. \footnote{https://christophm.github.io/interpretable-ml-book/shapley.html} So this method is a very good way of understanding contribution of each feature towards the prediction label for a data instance.
    \begin{figure}[!htbp]
  \begin{center}
    \includegraphics[width=8cm,height=7cm]{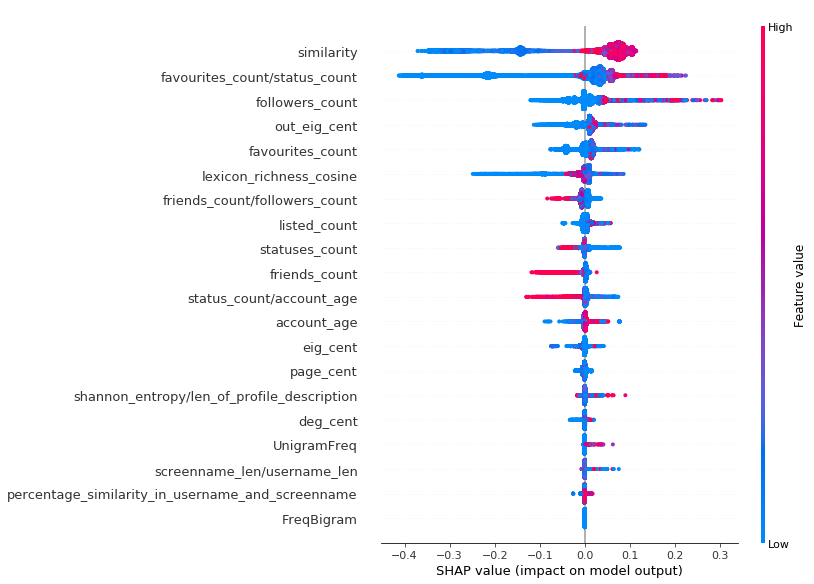}
    \caption{SHAP output for our feature set}
  \end{center}
\end{figure}

\subsubsection{\textbf{Correlation Matrix for all attributes \cite{guyon} } - \textit{Selection of all attributes equal or above a certain threshold - }}
 
    A relevant feature will always be highly correlated to the class and not redundant to any other relevant features. We use univariate feature analysis for selecting relevant features from the database and identifying redundant features and eliminating the same from original attribute set.
    \begin{figure}[!htbp]
  \begin{flushright}
    \includegraphics[width=10cm,height=9cm]{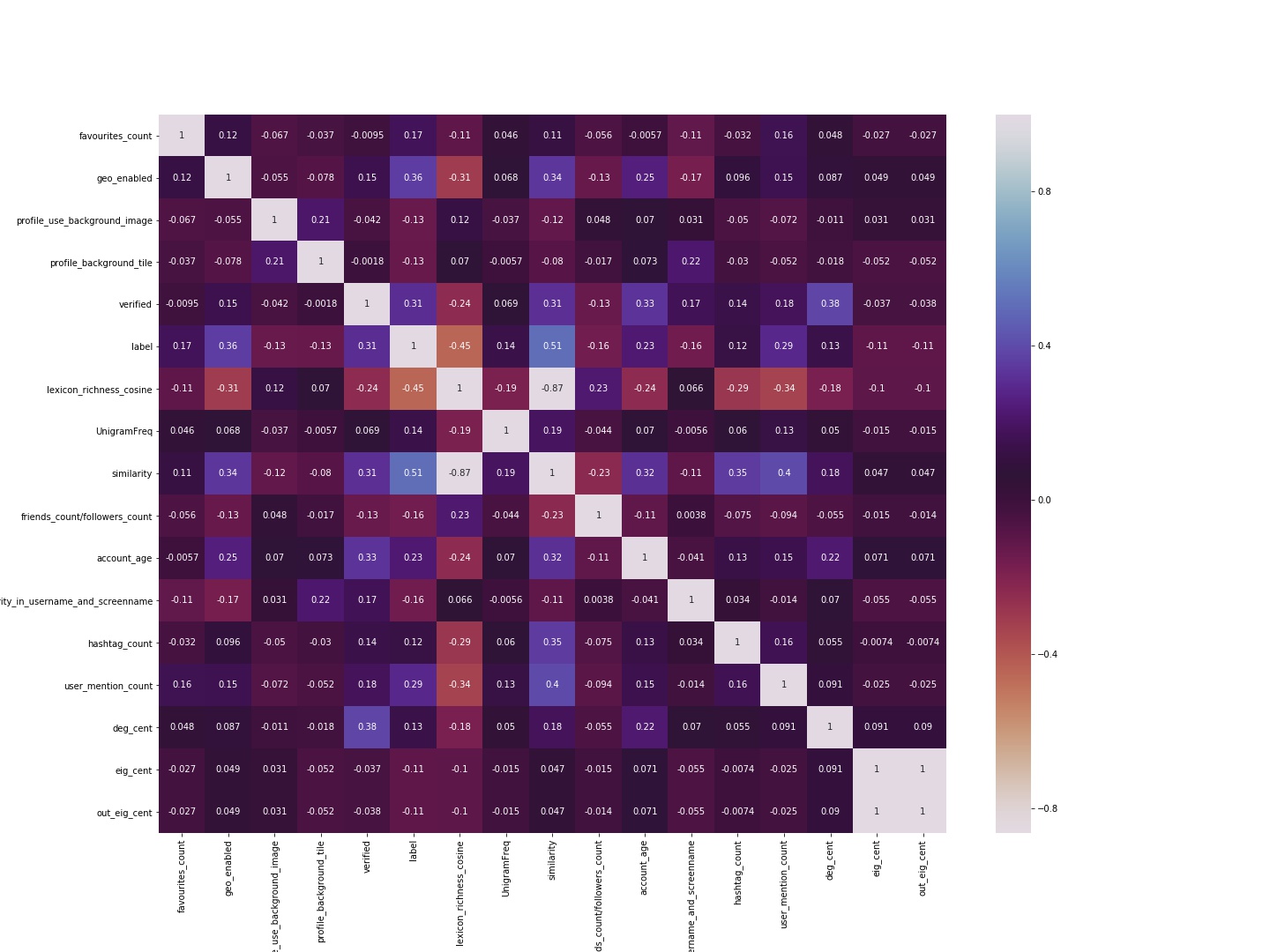}
    \caption{Visual heatmap of the univariate analysis of correlation for our feature set}
  \end{flushright}
\end{figure}
    
    Now we have a set $S1$ for top 15 features from SHAP and another set $S2$ for features with value equal or above 0.1 from the correlation matrix. Intersection of two sets $S1$ and $S2$, which consists of 16 features, are selected.
    The selected features are:
    \begin{multicols}{2}
    \begin{itemize}
        \item favourites\_count 
        \item geo\_enabled
        \item profile\_use\_background\_image 
        \item profile\_background\_tile 
        \item verified 
        \item lexicon\_richness\_cosine
         \item UnigramFreq 
         \item tweet\_similarity
         \item friends\_count/followers\_count 
         \item account\_age
         \item percentage\textunderscore similarity in user\-name and screenname
         \item hashtag\_count
         \item user\_mention\_count
         \item degree\_centrality
         \item in\_eig\_centrality 
         \item out\_eig\_centrality 
    \end{itemize}
    \end{multicols}
    Now, for each user, the vector with 16 feature values ($U_i$) and the 100 length embedding vector ($V_i$) generated from node2vec algorithm, are appended to create a vector of length 116 ($W_i$), which is our final attribute set for each Twitter user in our dataset. 
    
\section{Model Training and Evaluation}
We have tried a number of supervised Machine Learning algorithms to train our spam detection model. The details and corresponding results are given below -

\begin{multicols}{2}
\begin{itemize}
    \item Support Vector Machines - Polynomial and RBF Kernel (SVM)
    \cite{vapnik}
    \item Logistic Regression(LR)\cite{cramer}
    \item Naive Bayes (NB) \cite{rish}
    \item Neural Network - 3 layers (3NN) \cite{hinton}
    \item Random Forest(RF) \cite{ho}
    \item GBM and LightGBM \cite{ke}
    \item XGBoost (Decision Trees as Weak Classifiers) \cite{Chen}
\end{itemize}
\end{multicols}

\begin{center}
\begin{table}[!htbp]
    \caption{Accuracy on Class 0(\textit{Genuine} accounts), Class 1 (\textit{Spam} accounts), average accuracy and F1-scores of various machine learning models }
    \begin{tabular}{ p{1.5cm}p{1.5cm}p{1.5cm}p{1.5cm}p{1cm}  }
 
 \toprule
 \textbf{Model Name}& \textbf{Accuracy (Class 0)} & \textbf{Accuracy (Class 1)}&\textbf{Accuracy (Average)}&\textbf{F1-Score}\\
 \midrule
 SVM-Poly   & 79\%    &76\%&   77.62\% &0.77\\

 LR&   84\%  & 81\%   &82.49\% &0.83\\

 SVM-RBF &82\% & 80\%&  81\% &0.81\\

 3NN    &92\% & 90.62\%&  91\% &0.91\\

 RF&   95.22\%  & 94.75\%&94.98\% &0.94\\
 
 GBM& 90\%  & 87\%   &88.49\% &0.88\\

 \textbf{XGBoost}& \textbf{97.90}\%  & \textbf{96.08}\%&\textbf{96.99}\% &\textbf{0.97}\\
 \bottomrule
\end{tabular}
\end{table}
\end{center}

\begin{figure}[!htbp]
  \begin{center}
    \includegraphics[width=7cm,height=8cm,keepaspectratio]{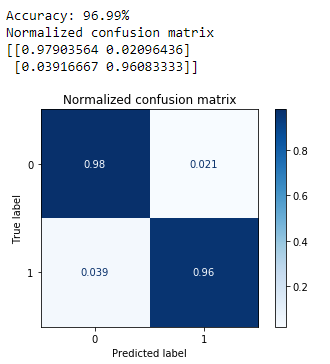}
    \caption{Detailed result of the XGBoost model}
  \end{center}
\end{figure}

Out of all the models that we have used, XGBoost gave significantly better performance than the other models as stated in Table 1. We trained our dataset with XGBoost ML Model, with Decision Trees as weak classifiers, Learning rate 0.1 and Max Depth of 15. These optimal parameters are found using Grid Search Cross Validation method. We now compare our model results with the same of previously stated works in section 2.

\begin{table}[!htbp]
    \caption{Comparison with other works }
    \begin{tabular}{ p{2.5cm}p{2cm}p{1.5cm}  }
 
 \toprule
 \textbf{Model}& \textbf{Dataset Size} & \textbf{Accuracy}\\
 \midrule
 Igawa et al.\cite{igawa}   & 100 & 96.6\%\\

 Varol et al.\cite{varol}   & 31K & 95\%\\

 Khaled et al.\cite{khaled} & 15K & 98.3\%\\

 Dutse et al.\cite{dutse}    & 2000 & 98.93\%\\

 Daouadi et al.\cite{daudi}   & 24K & 97.55\%\\
 
 Abu-El-Rub and Mueen\cite{abu} & 6 Million & 87.5\%\\

\textbf{Our(XGBoost)} & \textbf{46K} & \textbf{96.99\%}\\

 \bottomrule
\end{tabular}
\end{table}

We can see from Table 2 that, as the data size increases, the model performance decreases. Moreover, most of the models in Table 2 use databases of Twitter user IDs from 2018 and before, when there was a huge surge of fake,spam and bot accounts in Twitter. Only after mid 2018, Twitter brought robust spam detection and deletion technologies into their platform to get rid of most of them. \footnote{\url{https://blog.Twitter.com/en_us/topics/company/2018/how-Twitter-is-fighting-spam-and-malicious-automation.html}}. Now, we have 46k Twitter accounts in our database, which is second largest in Table 2. Significantly, all of these user accounts have successfully passed Twitter's spam detection tool to be remained active currently in the platform.\footnote{as of May 2020} So, these user accounts are much harder to be detected as spam accounts. Even though we have an updated and large database of such users in them, we successfully achieved an accuracy score of \~{}97\% from our model, which is a significant improvement over the baseline. The three other models that have higher accuracy scores than us, uses significantly smaller and much older Twitter database to train, consisting of more easily detectable fake/spam accounts.

\section{Conclusion and Future Work}

This work offers a novel and effective spam detection method that utilises an optimised set of user metadata attributes, features generated from NLP techniques and follower-following graph representation of user accounts. The effectiveness of the proposed features set is shown by a number of machine learning model evaluations and comparisons with previous state-of-the-art models. Performance is consistent across both class labels and different models. Also, there is significant improvement over the baseline. 

During data analysis, we observed that spam users tend to be selective in following other users, thereby forming communities of spammers. This is a high-level observation that we aim to explore further in the future. The clear difference between legitimate human users vs. legitimate social bots as well as human spammers vs. social bot spammers needs to be investigated further. Another interesting dimension for future work is to study the effect of the recent increase in the maximum length of tweets for posting, because it is true that automated spam accounts will face difficulties in generating lengthier tweets intelligently. Our Spam Account detection approach can also be applied in any ocial Media Platforms (SMPs) or real-time filtering application for data collection pipelines.


\bibliographystyle{ACM-Reference-Format}
\bibliography{sample-base}


\end{document}